\pdfoutput=1

\documentclass[11pt]{article}

\usepackage{acl2023}

\usepackage{times}
\usepackage{latexsym}

\usepackage{soul}

\usepackage[T1]{fontenc}

\usepackage[utf8]{inputenc}

\usepackage{microtype}

\usepackage{inconsolata}

\usepackage{amsmath}

\usepackage[noabbrev,capitalize]{cleveref}
\usepackage{multirow}
\usepackage{graphicx}
\usepackage{colortbl}
\usepackage[normalem]{ulem}

\usepackage[export]{adjustbox}

\usepackage{arydshln}
\def\hideXXX#1{}

\def\yesXXXifaccepted#1{#1}

\def\hidePV#1{}

\def\footurl#1{\footnote{\url{#1}}}

\def\inparcite#1{\citealp{#1}} 
\def\furl#1{\footnote{\url{#1}}}

\title{MT Metrics Correlate with Human Ratings of Simultaneous~Speech~Translation}

\author{Dominik Macháček$^1$ \and Ondřej Bojar$^1$ \and  Raj Dabre$^2$
  \\ \\
  Charles University, Faculty of Mathematics and Physics, \\ 
  Institute of Formal and Applied Linguistics$^{1}$ \\
   \\
  National Institute of Information and Communications Technology, Kyoto, Japan$^2$\\
  $^{1}$\texttt{\{machacek,bojar\}@ufal.mff.cuni.cz}, $^2$\texttt{raj.dabre@nict.go.jp}}

\def\chrf{chrF2}%

\def\system#1{#1}
\def\UPV{\system{UPV}}
\def\FBK{\system{FBK}}
\def\NAIST{\system{NAIST}}
\def\HWTSC{\system{HW-TSC}}
\def\CUNIKIT{\system{CUNI-KIT}}
\def\BertScore{\textsc{BertScore}}
\def\BertScoreShort{\textsc{BertS.}}
\begin{document}
\maketitle
\begin{abstract}
There have been several meta-evaluation studies on the correlation between human ratings and offline machine translation (MT) evaluation metrics such as BLEU, \chrf{},  \BertScore{} and COMET. These metrics have been used to evaluate simultaneous speech translation (SST) but their correlations with human ratings of SST, which has been recently collected as Continuous Ratings (CR), are unclear. In this paper, we leverage the evaluations of candidate systems submitted to the English-German SST task at IWSLT 2022 and conduct an extensive correlation analysis of CR and the aforementioned metrics. Our study reveals that the offline metrics are well correlated with CR and can be reliably used for evaluating machine translation in simultaneous mode, with some limitations on the test set size. We conclude that given the current quality levels of SST, these metrics can be used as proxies for CR, alleviating the need for large scale human evaluation. 
Additionally, we observe that correlations of the metrics with translation as a reference is significantly higher than with simultaneous interpreting, and thus we recommend the former for reliable evaluation.
\end{abstract}


\section{Introduction}




The current approach to evaluate simultaneous speech translation (SST, \inparcite{cho-esipova,ma-etal-2019-stacl}) systems that
have text as the output modality is to use automatic  metrics which are designed for offline text-to-text machine translation (MT), alongside other measures for latency and stability.
Researchers tend to use offline metrics, such as BLEU \cite{papineni-etal-2002-bleu}, \chrf{} \cite{popovic-2017-chrf}, \BertScore{} \cite{bert-score}, COMET \cite{rei-etal-2020-unbabels} and others \cite{freitag-EtAl:2022:WMT} in SST despite no explicit evidence that they correlate with human ratings. 



However, simultaneous speech-to-text translation has different characteristics compared to offline text-to-text MT. For example, when the users are following subtitles in real-time, they have limited time for reading and comprehension as they cannot fully control the reading pace by themselves. Therefore, they may be less sensitive to subtle grammar and factual flaws than while reading a text document without any time constraints. The human evaluation of SST should therefore reflect the simultaneity. The users may also prefer brevity and simplicity over verbatim word-for-word translation. Even if the reference is brief and simpler than the original, there may be lots of variants that the BLEU score and other MT metrics may not evaluate as correct. 

\yesXXXifaccepted{Furthermore, SST and MT differ in their input modalities. MT sources are assumed to originate as texts, while the SST source is a speech given in a certain situation, accompanied by para-linguistic means and specific context knowledge shared by the speaker and listener. Transcribing speech to text for use in offline evaluation of SST may be limiting.}



In this paper, we aim to determine the suitability of automatic metrics for evaluating SST.
To this end, we analyze the results of the simultaneous speech translation task from English to German at IWSLT 2022 \cite{anastasopoulos-etal-2022-findings}, where we calculate the correlations between MT metrics and human judgements in simultaneous mode. There are five competing systems and human interpreting that are manually rated by bilingual judges in a simulated real-time event. Our studies show that BLEU does indeed correlate with human judgements of simultaneous translations under the same conditions as in offline text-to-text MT: on a sufficiently large number of sentences. Furthermore, \chrf{}, \BertScore{} and COMET exhibit similar but significantly larger correlations. To the best of our knowledge, we are the first to explicitly establish the correlation between automatic offline metrics with human SST ratings, indicating that they may be safely used in SST evaluation in the currently achieved translation quality levels.

Additionally, we statistically compare the metrics with translation versus interpreting reference, and we recommend the most correlating one: translation reference and COMET metric, with \BertScore{} and \chrf{} as fallback options.


We publish the code for analysis and visualisations that we created in this study.\furl{github.com/ufal/MT-metrics-in-SimST}
It is available for further analysis and future work.

\section{Related Work}

We replicate the approach from text-to-text MT research (e.g.\ \inparcite{papineni-etal-2002-bleu}) that examined the correlation of MT metrics with human judgements. The strong correlation is used as the basis for taking the metrics as reliable. As far as we know, we are the first who apply this approach to SST evaluation in simultaneous mode. 


In this paper, we analyze four metrics that represent the currently used or recommended \cite{freitag-EtAl:2022:WMT} types of MT metrics. BLEU and \chrf{} are based on lexical overlap and are available for any language. \BertScore{} \cite{bert-score} is based on embedding similarity of a pre-trained BERT language model. COMET \cite{rei-etal-2020-unbabels} is a neural metric trained to estimate the style of human evaluation called Direct Assessment \cite{graham-etal-2015-accurate}. COMET requires sentence-to-sentence aligned source, translation and reference in the form of texts, which may be unavailable in some SST use-cases; then, other metric types may be useful. Another fact is that \BertScore{} and COMET are available only for a limited set of  languages. 








\section{Human Ratings in SST}



As far as we know, the only publicly available collection of simultaneous (not offline) human evaluation of SST originates from IWSLT 2022 \cite{iwslt-2022-international} English-to-German Simultaneous Translation Task, which is described in ``Findings'' (\inparcite{anastasopoulos-etal-2022-findings}, see highlights of it we discuss in \Cref{sec:highlights}). The task focused on speech-to-text translation and was reduced to translation of individual sentences. The segmentation of the source audio to sentences was provided by organizers, and not by the systems themselves. 
The source sentence segmentation that was used in human evaluation was gold (oracle). 
\yesXXXifaccepted{It only approximates a realistic setup where the segmentation would be provided by an automatic system, e.g.\ \citet{tsiamas22_interspeech}, and may be partially incorrect and cause more translation errors compared to the gold segmentation.}

\yesXXXifaccepted{The simultaneous mode in Simultaneous Translation Task means that the source is provided gradually, one audio chunk at a time. After receiving each chunk, the system decides to either wait for more source context, or produce target tokens. Once the target tokens are generated, they can not be rewritten.}

The participating systems are submitted and studied in three latency regimes: low, medium and high. It means that the maximum Average Lagging \cite{ma-etal-2019-stacl} between the source and target on validation set must be 1, 2 or 4 seconds in a ``computationally unaware'' simulation where the time spent by computation, and not by waiting for context, is not counted. \yesXXXifaccepted{One system in low latency did not pass the latency constraints (see Findings, page 44, numbered 141), but it is manually evaluated regardless.} 

Computationally unaware latency was one of the main criteria in IWSLT 2022. It means that the participants did not need to focus on a low latency implementation, as it is more of a technical and hardware issue than a research task. However, the subtitle timing in manual evaluation was created in a way such that waiting for the first target token was dropped, and then it continued with computationally aware latency.

\subsection{Continuous Rating (CR)}

Continuous Rating (CR, \inparcite{ContinuousRating2022,presenting-machacek-bojar-2020}) is a method for human assessment of SST quality in a simulated online event. An evaluator with knowledge of the source and target languages watches a video (or listens to an audio) document with subtitles created by the SST system which is being evaluated. The evaluator is asked to continuously rate the quality of the translation by pressing buttons with values 1 (the worst) to 4 (the best). Each evaluator can see every document only once, to ensure one-pass access to the documents, as in a realistic setup. 

CR is analogous to Direct Assessment \cite{graham-etal-2015-accurate}, which is a method of human text-to-text MT evaluation in which a bilingual evaluator expresses the MT quality by a number on a scale. It is natural that individual evaluators have different opinions, and thus it is a common practice to have multiple evaluators evaluate the same outputs and then report the mean and standard deviation of evaluation scores, or the results of statistical significance tests that compare the pairs of candidate systems and show how confident the results are.

\citet{ContinuousRating2022} showed that CR relates well to comprehension of foreign language documents by SST users. Using CR alleviates the need to evaluate comprehension by factual questionnaires that are difficult to prepare, collect and evaluate. Furthermore, \citet{ContinuousRating2022} show that bilingual evaluators are reliable.


\paragraph{Criteria of CR}
In IWSLT 2022, the evaluators were instructed that the primary criterion in CR should be meaning preservation (or adequacy), and other aspects such as fluency should be secondary. The instructions do not mention readability due to output segmentation frequency or verbalizing non-linguistic sounds such as ``laughter'', despite the system candidates differ in these aspects. 

\subsection{Candidate Systems}

\paragraph{Automatic SST systems}
There are 5 evaluated SST systems:
\FBK{} \cite{gaido-etal-2022-efficient}, \NAIST{} \cite{fukuda-etal-2022-naist}, \UPV{} \cite{iranzo-sanchez-etal-2022-mllp}, \HWTSC{} \cite{wang-etal-2022-hw-tscs}, and \CUNIKIT{} \cite{polak-etal-2022-cuni}. 

\paragraph{Human Interpreting} In order to compare the state-of-the-art SST with human reference, the organizers hired one expert human interpreter to simultaneously interpret all the test documents. Then, they employed annotators to transcribe the voice into texts. The annotators worked in offline mode. The transcripts were then formed as subtitles including the original interpreter's timing and were used in CR evaluation the same way as SST.
However, human interpreters use their own segmentation to translation units so that they often do not translate one source sentence as one target sentence. There is no gold alignment of the translation sentences to interpreting chunks. The alignment has to be resolved before applying metrics to interpreting.


\subsection{Evaluation Data}

There are two subsets of evaluation data used in IWSLT22 En-De Simultaneous Translation task. The ``Common'' subset consists of TED talks of the native speakers.\yesXXXifaccepted{See the description in Findings on page 9 (numbered as 106).} 
The ``Non-Native'' subset consists of mock business presentations of European high school students \cite{antrecorp-paper}, and of presentations by representatives of European supreme audit institutions. 
\yesXXXifaccepted{This subset is described in Findings on page 39 (numbered page 136). The duration statistics of audio documents in both test sets are in Findings in Table~17 on page 48 (numbered 145).}



\section{Correlation of CR and MT Metrics}
\label{sec:correlation}

In this section, we study the correlation of CR and MT metrics BLEU, \chrf{}, \BertScore{} and COMET.  
We measure it on the level of documents, and not on the test set level, increasing the number of observations for significance tests. There are 60 evaluated documents (17 in the Common subset and 43 in Non-Native) and 15 system candidates (5 systems, each in 3 latency regimes), which yields 900 data points. 

We discovered that \CUNIKIT{} system outputs are tokenized, while the others are detokenized. Therefore, we first detokenized \CUNIKIT{} outputs.
Then, we removed the final end of sequence token (\texttt{</s>}) from the outputs of all systems. 
Finally, we calculated BLEU and \chrf{} using sacreBLEU \cite{post-2018-call}, \BertScore{} and COMET. See \Cref{sec:metric-sign} for metric details and signatures. 

\yesXXXifaccepted{In total, there are 1584 rating sessions of 900 candidate document translations. Each candidate document translation is rated either twice with different evaluators, once, or not at all. 
}
We aggregate the individual rating clicks in each rating session by plain average (CR definition in \Cref{sec:aggregating}) to get the CR scores. Then, we average the CR of the same documents and candidate translations, and we correlate it with MT metrics.

   \def\hfigcorr{0.49\textwidth}
\begin{figure}[th!]
    \centering
 \includegraphics[width=\hfigcorr]{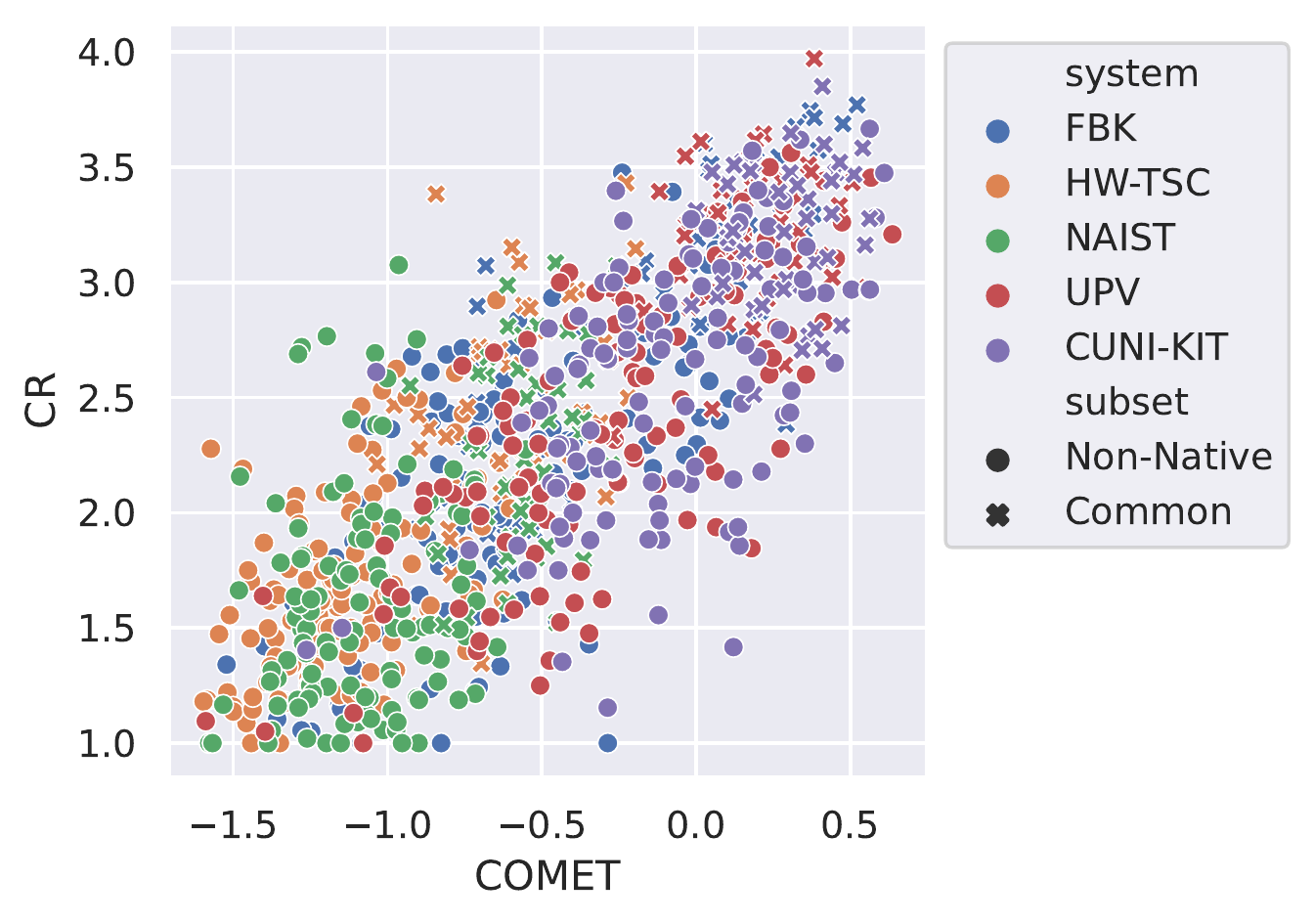}
    \caption{Averaged document CR vs COMET on both Common and Non-Native subsets.}
    \label{fig:CRcorr}
\end{figure}

\def\nocorr#1{\textcolor{gray}{#1}}
\def\nocorr#1{\textit{#1}}
\begin{table}[h!]
    \centering
    \footnotesize
    \begin{tabular}{l|r|r@{~~}r@{~~}r@{~~}r@{~~}r}
    \multicolumn{6}{c}{\textbf{Averaged document ratings}}   \\
 subsets        & num. &  BLEU  & \chrf{}&\BertScoreShort{}&COMET\\
 \hline
 both           & 823 & 0.65  &    0.73 &  0.77 & 0.80 \\
 Common & 228 & \nocorr{0.42} &     0.63 & 0.68 & 0.76 \\
 Non-Native    & 595 & 0.70   &   0.70 &  0.73 & 0.75 \\
 \multicolumn{6}{c}{} \\
     \multicolumn{6}{c}{\textbf{All document ratings}}   \\
subsets        & num. &  BLEU  &\chrf{}&\BertScoreShort{}&COMET\\
 \hline
 both           & 1584 &  0.61  &  0.68 &  0.71 & 0.73 \\
 Common & 441 & \nocorr{0.37}   &   \nocorr{0.57} & 0.60 & 0.68 \\
 Non-Native    & 1143 &  0.64   &   0.64 & 0.66 & 0.67 \\
    \end{tabular}
    \caption{Pearson correlation coefficients for CR vs MT metrics BLEU, \chrf{}, \BertScore{} and COMET for averaged document ratings by all 5 SST systems and 3 latency regimes (upper), and all ratings (lower). When the coefficient is less than 0.6 (in \nocorr{italics}), 
    the correlation is not considered as strong. Significance values are $p<0.01$ in all cases, meaning strong confidence.}
    \label{tab:CRcorr}
\end{table}

\paragraph{Correlation Results} In \Cref{tab:CRcorr}, we report correlation coefficients with and without averaging, together with the number of observations. \Cref{fig:CRcorr} displays the relation between CR and COMET.

Pearson correlation is considered as strong if the coefficient is larger than 0.6 \cite{evans-1996}. The results show strong correlation (above 0.65) of CR with BLEU, \chrf{}, \BertScore{} and COMET at the document level on both test subsets. When we consider only one subset, the correlation is lower, but still strong for \chrf{}, \BertScore{} and COMET (0.63, 0.68 and 0.76, resp.). It is because the Common subset is generally translated better than Non-Native, so with only one subset, the points span a smaller part of the axes and contain a larger proportion of outliers.

The strong correlation is not the case of BLEU on the Common subset where the Pearson coefficient is 0.42. We assume it is because BLEU is designed for use on a larger test set, but we use it on short single documents. However, BLEU  correlates with \chrf{} and COMET (0.81 and 0.62 on the Common subset). 
BLEU also correlates with CR on the level of test sets, as reported in Findings in the caption of Table~18 (page 48, numbered 145). 

We conclude that \ul{with the current overall levels of speech translation quality, BLEU, \chrf{}, \BertScore{} and COMET can be used for reliable assessment of human judgement of SST quality at least on the level of test sets. \chrf{}, \BertScore{} and COMET are reliable also at the document level}.

\def\singleseq{\textsc{SingleSeq}}
\def\ulsingleseq{\textsc{\ul{SingleSeq}}}
\def\sent{\textsc{Sent}}
\def\mwer{\textsc{mWER}}
\def\transl{\textsc{transl}}
\def\intp{\textsc{intp}}

\begin{table}[t]
    \centering
    \footnotesize
    \begin{tabular}{lll|r}
\textbf{metric} & \textbf{reference} & \textbf{alignment} & \textbf{corr.} \\
\hline
\hline
COMET & \transl{} & \sent{} & 0.80 \\
COMET & \transl{} & \singleseq & 0.79 \\
COMET & \transl{}+\intp{} & \singleseq & 0.79 \\
\hdashline
\BertScore{} & \transl{} & \sent{} & 0.77 \\
\BertScore{} & \transl{}+\intp{} & \sent{}+\mwer{} & 0.77 \\
COMET & \intp{} & \singleseq & 0.77 \\
\BertScore{} & \transl{}+\intp{} & \singleseq & 0.76 \\
\BertScore{} & \transl{} & \singleseq & 0.75 \\
\hdashline[1pt/1pt]
\chrf{} & \transl{}+\intp{} & \sent{}+\mwer{} & 0.73 \\
BLEU & \transl{}+\intp{} & \singleseq & 0.73 \\
\chrf{} & \transl{} & \sent{} & 0.73 \\
\chrf{} & \transl{}+\intp{} & \singleseq & 0.72 \\
\chrf{} & \transl{} & \singleseq & 0.72 \\
BLEU & \transl{} & \singleseq & 0.71 \\
COMET & \intp{} & \mwer{} & 0.71 \\
\BertScore{} & \intp{} & \singleseq & 0.69 \\
BLEU & \transl{}+\intp{} & \sent{}+\mwer{} & 0.68 \\
\chrf{} & \intp{} & \singleseq & 0.66 \\
BLEU & \transl{} & \sent{} & 0.65 \\
\chrf{} & \intp{} & \mwer{} & 0.65 \\
BLEU & \intp{} & \singleseq & 0.65 \\
\hdashline
\BertScore{} & \intp{} & \mwer{} & 0.60 \\
BLEU & \intp{} & \mwer{} & 0.58 \\
    \end{tabular}
    \caption{Pearson correlation of metric variants to averaged CR on both subsets, ordered from the most to the least correlating ones. Lines indicate ``clusters of significance'', i.e. boundaries between groups where all metric variants significantly differ from all in the other groups, with $p<0.05$ for dashed line and $p<0.1$ for dotted line. See the complete pair-wise comparison in \Cref{sec:heatmaps}.}
    \label{tab:metrics-variants}
\end{table}

\paragraph{Translation vs Interpreting Reference}
There is an open question whether SST should rather mimic offline translation, or simultaneous interpreting. As \citet{machacek21_interspeech} discovered, translation may be more faithful, word-for-word, but also more complex to perceive by target audience. Simultaneous interpreting, on the other hand, tends to be brief and simpler than offline translation. However, it may be less fluent and less accurate. 
Therefore, we consider human translation (\transl{}) and transcript of simultaneous interpreting (\intp{}) as two possible references, and also test multi-reference metrics with both.

Since interpreting is not sentence-aligned to SST candidate translations, we consider two alignment methods: single sequence (\singleseq{}), and mWERSegmenter (\inparcite{matusov-etal-2005-evaluating}, \mwer{}). 
\singleseq{} method means that we concatenate all the sentences in the document to one single sequence, and then apply the metric on it, as if it was one sentence. mWERSegmenter is a tool for aligning translation candidates to reference, if their sentence segmentation differs. It finds the alignment with the minimum WER when comparing tokens in aligned segments. 
For translation, we also apply the default sentence alignment (\sent{}).

In \Cref{tab:metrics-variants}, we report the correlations of metric, reference and alignment variants and their significance, with more details in \Cref{sec:heatmaps}.

\subsection{Recommendations}
Taking CR as the golden truth of human quality, we make the following recommendations of the most correlating metric, reference and sentence alignment method for SST evaluation. 

\paragraph{Which metric?} \ul{COMET}, because it correlates significantly better with CR than \BertScore{} does. From the fall back options, \chrf{} should be slightly 
preferred over BLEU.

\paragraph{Which reference?} The metrics give significantly higher correlations with CR with translations than with interpreting as a reference. Difference between translation reference and two references (\transl{}+\intp{}) is insignificant. Therefore, we recommend \ul{translation as a reference for SST}. 

\paragraph{Which alignment method?} With an unaligned reference, COMET and \BertScore{} correlate significantly more with \ulsingleseq{} than with \mwer{}, probably because the neural metrics are trained on full, complete sentences, which are often split to multiple segments by mWERSegmenter. \chrf{} correlates insignificantly better with \underline{\mwer{}} than with \singleseq{}. 

\section{Conclusion}

We found correlation of offline MT metrics to human judgements of simultaneous speech translation. The most correlating and thus preferred metric is COMET, followed by \BertScore{} and \chrf{}. We recommend text translation reference over interpreting, and single sequence alignment for neural, and mWERSegmenter for $n$-gram metrics.



\section{Limitations}

The data that we analyzed are limited to only one English-German language pair, 5 SST systems from IWSLT 2022, and three domains. All the systems were trained in the standard supervised fashion on parallel texts. They do not aim to mimic interpretation with shortening, summarization or redundancy reduction, and they do not use document context. The used MT metrics are good for evaluating individual sentence translations and that is an important, but not the only subtask of SST. We assume that some future systems created with a different approach may show divergence of CR and the offline MT metrics.

Furthermore, we used only one example of human interpreting. A precise in-depth study of human interpretations is needed to re-assess the recommendation of  translation or interpreting as reference in SST.


\section*{Acknowledgements}

We are thankful to Dávid Javorský and Peter Polák for their reviews. 


This research was partially supported by the grants 19-26934X  (NEUREM3)  of  the  Czech  Science Foundation, 
SVV project number 260~698, and 398120 of the Grant Agency of Charles University.

\bibliography{anthology,custom}
\bibliographystyle{acl_natbib}


\appendix

\section{Highlights of IWSLT22 Findings} 
\label{sec:highlights}

\begin{table*}[ht!]
    \centering
    \begin{tabular}{l|cc|l}
marker     &  PDF page & numbered page & description \\
\hline
Section 2 & 3-5 & 100-102 & Simultaneous Speech Translation Task \\
Figure 1 & 6 & 103 & Quality-latency trade-off curves \\
Section 2.6.1         &  5 & 102 & Description of human evaluation \\
Figure 5    & 8 & 105 & Manual scores vs BLEU (plot)\\
Two Test Sets (paragraph) & 39 & 136 & Non-Native subset \\
Test data (paragraph) & 9 & 106 & Common (native) subset of test data \\
Automatic Evaluation Results & 44 & 141 & Latency and BLEU results (table) \\
A1.1 (appendix)  &   38-39 & 135-136 & Details on human evaluation \\
Table 17 & 48 & 145 & Test subsets duration \\
Table 18 & 48 & 145 & Manual scores and BLEU (table) \\ 
    \end{tabular}
    \caption{Relevant parts of IWSLT22 Findings (\url{https://aclanthology.org/2022.iwslt-1.10v2.pdf}) for En-De Simultaneous Speech Translation task and human evaluation.}
    \label{tab:findings-parts}
\end{table*}

The Findings of IWSLT22 \cite{anastasopoulos-etal-2022-findings} are available in PDF. The most up-to-date version (version 2) is 61 pages long.\furl{https://aclanthology.org/2022.iwslt-1.10v2.pdf} We highlight the relevant parts of Findings with page numbers in \Cref{tab:findings-parts} so that we can refer to them easily.

Note that findings are a part of the conference proceedings \cite{iwslt-2022-international} as a chapter in a book. The order of findings pages in PDF does not match the page numbers at the footers.

Also note that in Section 2.4 on page 4 (in PDF, 101 in Proceedings), there is a description of \textsc{MLLP-VRAIN} which corresponds to the system denoted as \UPV{} in all other tables and figures.

\section{Metric Signatures}
\label{sec:metric-sign}

BLEU and \chrf{} SacreBLEU metric signature is {case:mixed|eff:no|tok:13a|smooth:exp|version:2.3.1}.

For \BertScore{}, we used F1 with signature {{bert-base-multilingual-cased\_L9\_no-idf\_ver\-sion=0.3.12(hug\_trans=4.23.1)\_fast-tokenizer}}.

We use COMET model \texttt{wmt20-comet-da} \cite{rei-etal-2020-unbabels}. 
For multi-reference COMET, we run the model separately with each reference and average the scores.

The standard way of using mWERSegmenter is to segment candidate translation according to reference. However, COMET requires aligned source as one of the inputs, and mWERSegmenter can not align it because it is in other language. For COMET \intp{} \mwer{} variant, we therefore aligned interpreting to translation, which is already aligned to source. For the other metrics with \intp{} \mwer{}, we aligned translation candidate to interpreting, which is the standard way. 

\section{Aggregating Continuous Ratings}
\label{sec:aggregating}

We revisited the processing of the individual collected clicks on the rating buttons into the aggregate score of Continuous Rating. 


We found two definitions that can yield different results in certain situations: 
(1) The rating (as clicked by the evaluator) is valid at the instant time point when the evaluator clicked the rating button. The final score is the average of all clicks, each click has the equal weight. We denote this interpretation as $CR$.

(2) The rating is assigned to the time interval from the click time to the next click, or between the last click and the end of the document. The length of the interval is considered in averaging. The final score is the average of ratings weighted by interval lengths when the rating is valid. We denote this interpretation as $CRi$.
\footnote{Other interpretations are also conceivable, for instance assuming that the rating applies to a certain time before the click and then till the next judgement.}

To express them rigorously, let us have a document of duration $T$, and $n$ ratings $(r_i,t_i)$, where $i\in\{1,\dots,n\}$ is an index, $r_i \in \{1,\dots,4\}$ is the rated value and $0 \le t_1 < \dots < t_n \le T$ are times when the ratings were recorded. 

Then, the definitions are as follows:

$$CR = {1 \over n} \sum_{i=1}^n r_i$$


$$CRi = {1\over T-t_1} \Big(\sum_{i=1}^{n-1}(t{_{i+1}}-{t_i}) r_i + (T-t_n) r_n \Big)$$

If the judges press the rating buttons regularly, with a uniform frequency, then both definitions give equal scores. Otherwise, the $CR$ and $CRi$ may differ and may yield even opposite conclusions. For example, pressing ``1'' twelve times in one minute, then ``4'' and then waiting for one minute results in different scores: $CR=1.2$, $CRi = 2$.

\begin{figure}[ht]
    \centering
    \includegraphics[width=0.48\textwidth]{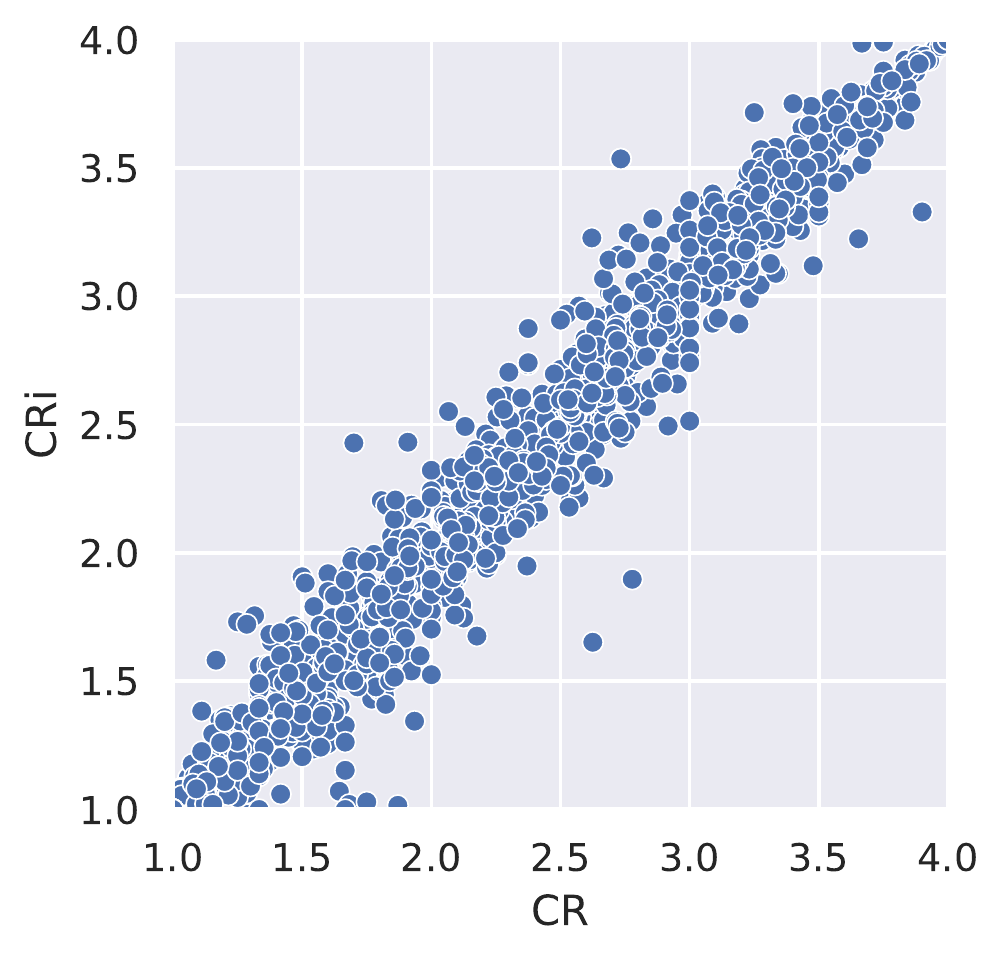}
    \caption{Relation between weighted interval averaging of continuous rating (CRi, y-axis) and average of all ratings (CR, x-axis) for each annotation of each document (blue data points).}
    \label{fig:cr_relation}
\end{figure}

To examine the relationship between these definitions, we counted $CR$ and $CRi$ for each annotation of each document in the evaluation campaign. The results are in \Cref{fig:cr_relation} where we observe correlation between the two definitions. The Pearson correlation coefficient is 0.98, which indicates a very strong correlation.



\paragraph{Summary}
Based on the correlation score we observed, we conclude that both definitions are interchangeable, and any of them can be used in further analysis.



\vskip 5em

\section{Pairwise Metrics Comparison}
\label{sec:heatmaps}

We test the statistical difference of correlations with Steiger's
method.\footnote{\url{https://github.com/psinger/CorrelationStats/}} The
method takes into account the number of data points and the fact that all
three compared variables correlate, which is the case of the MT metrics that
are applied on the same texts. We use two-tailed test. 

\begin{figure*}
    \centering
    \textbf{Both subsets}
    \includegraphics[width=\textwidth]{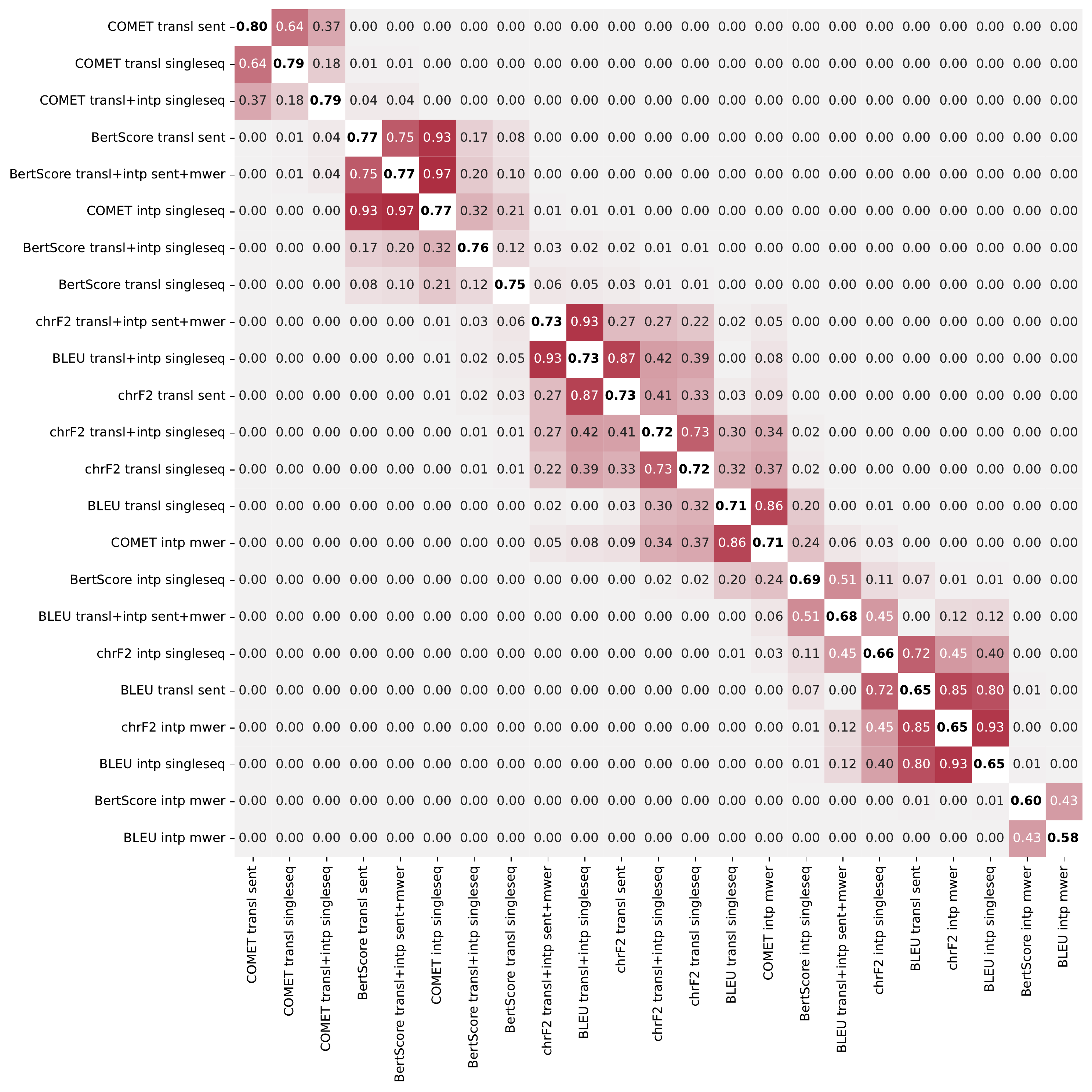}
    \caption{Results of significance test ($p$-values rounded to two decimal digits) for difference of correlations of the metrics variants to CR. The metrics variants are ordered by Pearson correlation to CR on both subsets from most correlating (top left) to least (bottom right). The bold numbers on the diagonal are the correlation coefficients to CR.}
    \label{fig:heatmap-both}
\end{figure*}

\begin{figure*}
    \centering
    \textbf{Common subset}
    \includegraphics[width=\textwidth]{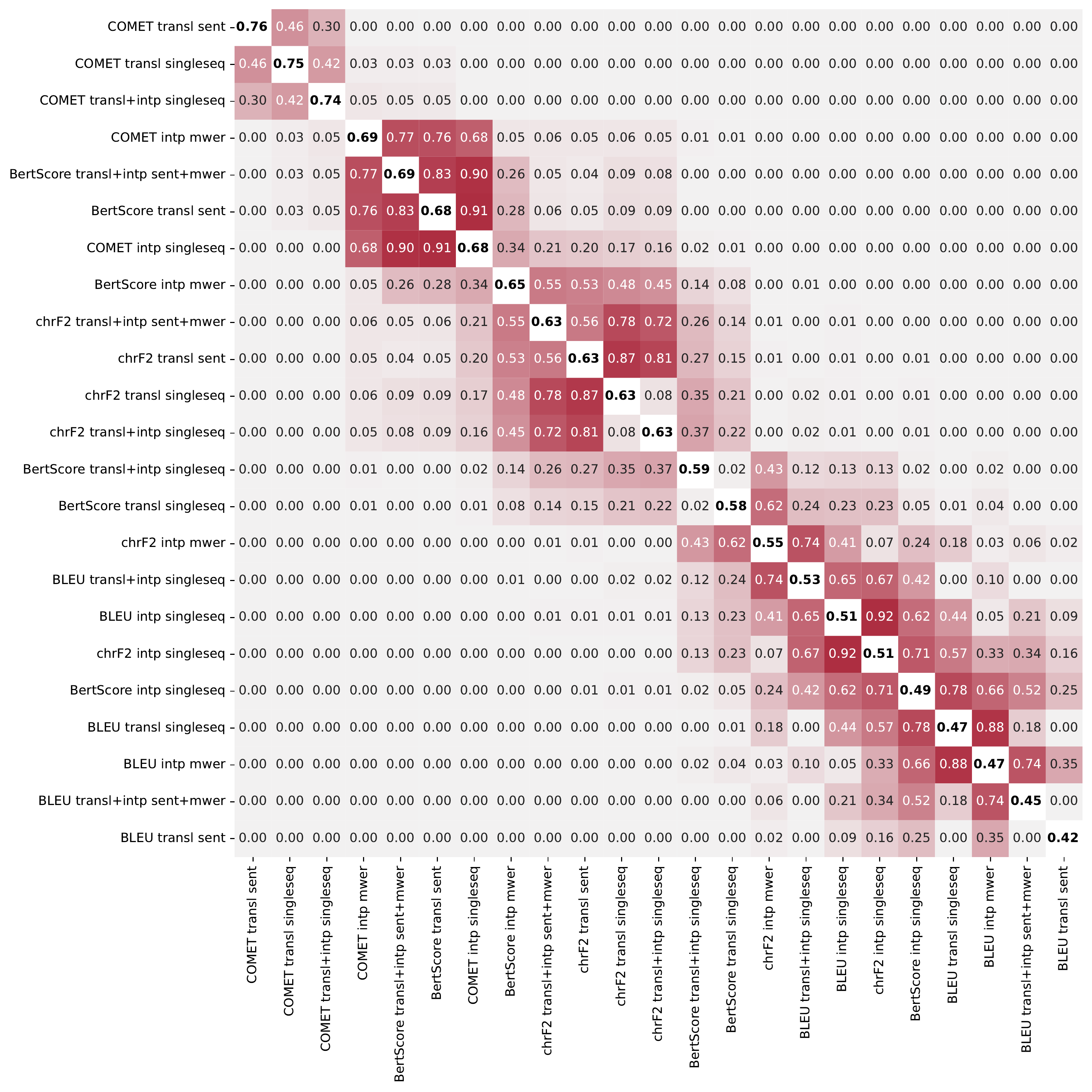}
    \caption{Results of significance test ($p$-values rounded to two decimal
	digits) for difference of correlations of the metrics variants to CR.
	The metrics variants are ordered by Pearson correlation to CR on the Common subset from most correlating (top left) to least (bottom right). The bold numbers on the diagonal are the correlation coefficients to CR.}
    \label{fig:heatmap-nat}
\end{figure*}

\begin{figure*}
    \centering
    \textbf{Non-Native subset}
    \includegraphics[width=\textwidth]{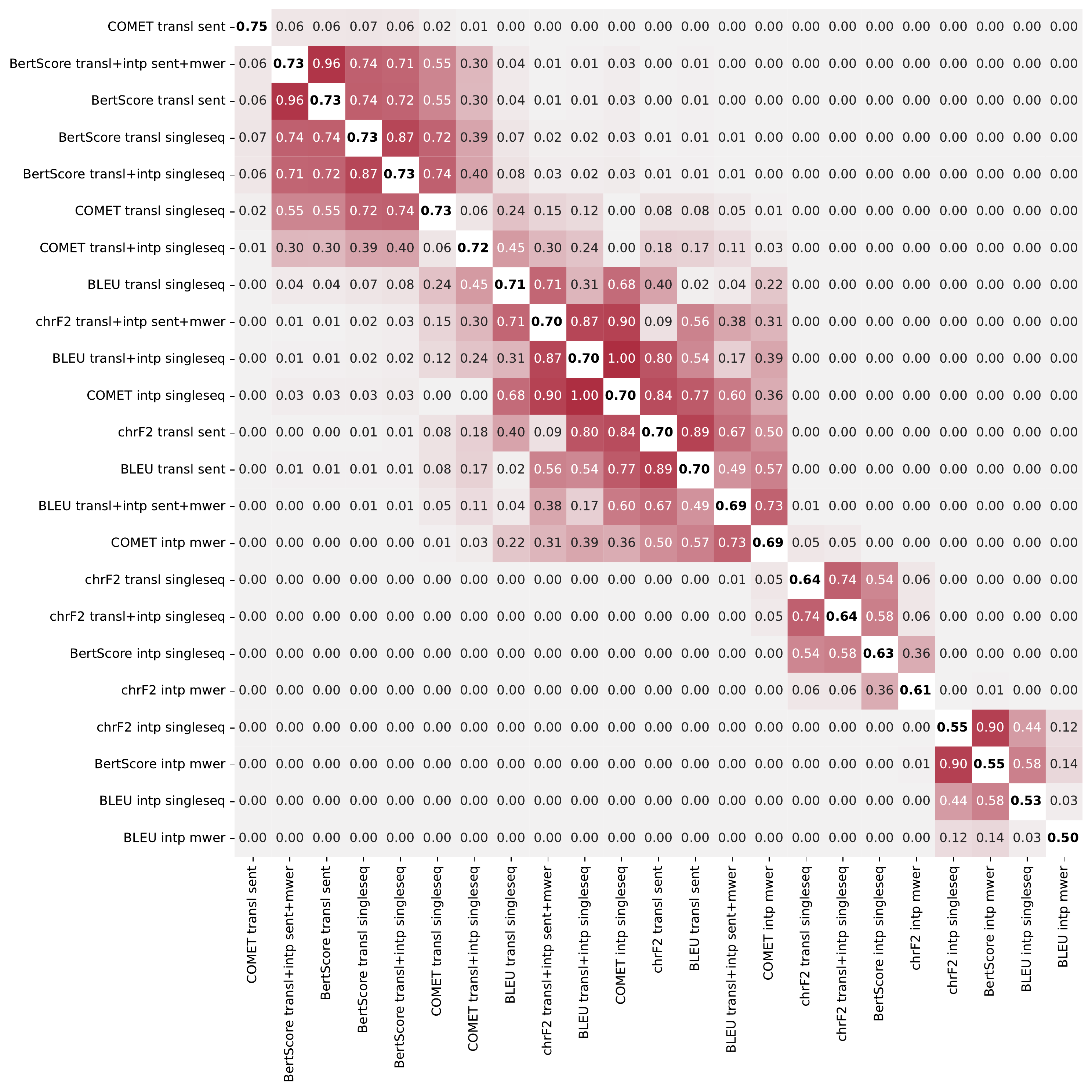}
    \caption{Results of significance test ($p$-values rounded to two decimal
	digits) for difference of correlations of the metrics variants to CR.
	The metrics variants are ordered by Pearson correlation to CR on the Non-Native subset from most correlating (top left) to least (bottom right). The bold numbers on the diagonal are the correlation coefficients to CR.}
    \label{fig:heatmap-nonn}
\end{figure*}

We applied the test on all pairs of metric variants. The results for both subsets are in \Cref{fig:heatmap-both}.  
\Cref{fig:heatmap-nat} displays results on the Common subset, and
\Cref{fig:heatmap-nonn} for the Non-Native subset. These results are analogous to
those in \Cref{tab:CRcorr} in \Cref{sec:correlation}. The correlation scores
for the two subsets treated separately
are lower and the differences along the diagonal are less significant. We
explain it by the fact that in smaller data set, there is larger impact of
noise.

\end{document}